\pdfoutput=1

\documentclass[11pt]{article}

\usepackage{graphicx}
\usepackage{amsmath}
\usepackage{booktabs}
\usepackage{multirow}
\usepackage{multicol}
\usepackage{stfloats}
\usepackage{bm}
\usepackage{ACL2023}

\usepackage{times}
\usepackage{latexsym}

\usepackage[T1]{fontenc}

\usepackage[utf8]{inputenc}

\usepackage{microtype}

\usepackage{inconsolata}

\title{NAPG: Non-Autoregressive Program Generation for Hybrid Tabular-Textual Question Answering}

\author{Tengxun Zhang$^{1}$\ \ \ \ Hongfei Xu$^{1}$\thanks{\ \ \ \ Corresponding author.}\ \ \ \ Josef van Genabith$^2$\ \ \ \ Deyi Xiong$^{3}$\ \ \ \ Hongying Zan$^{1}$\\
$^1$Zhengzhou University, Henan, China \\
$^2$DFKI and Saarland University, Informatics Campus, Saarland, Germany\\
$^3$Tianjin University, Tianjin, China\\
\{ztx313,hfxunlp\}@foxmail.com,
Josef.Van\_Genabith@dfki.de,\\
dyxiong@tju.edu.cn,
iehyzan@zzu.edu.cn}

\begin{document}
\maketitle
\begin{abstract}
Hybrid tabular-textual question answering (QA) requires reasoning from heterogeneous information, and the types of reasoning are mainly divided into numerical reasoning and span extraction. Current numerical reasoning methods autoregressively decode program sequences, and each decoding step produces either an operator or an operand. However, the step-by-step decoding suffers from exposure bias, and the accuracy of program generation drops sharply as the decoding steps unfold due to error propagation. In this paper, we propose a non-autoregressive program generation framework, which independently generates complete program tuples containing both operators and operands, can address the error propagation issue while significantly boosting the speed of program generation. Experiments on the ConvFinQA and MultiHiertt datasets show that our non-autoregressive program generation method can bring about substantial improvements over the strong FinQANet ($+5.06$ Exe Acc and $+4.80$ Prog Acc points) and MT2Net ($+7.97$ EM and $+6.38$ F1 points) baselines, establishing the new state-of-the-art performance, while being much faster ($\sim$21x) in program generation. Finally, with increasing numbers of numerical reasoning steps the performance drop of our method is significantly smaller than that of the baselines. Our code will be publicly available soon.
\end{abstract}

\section{Introduction}

Most previous QA studies focus on homogeneous data, such as unstructured text \citep{hermann2015teaching,chen2017reading,yang2018hotpotqa,li2020molweni,nie2020large} or structured knowledge bases \citep{yih2015semantic,weston2015towards,talmor2018web,zhang2020summarizing,zhang2020web}. In comparison, hybrid tabular-textual QA \citep{chen2020open,chen2020hybridqa,zhu2021tat,chen2021finqa,li2022learning,chen2022convfinqa,zhao2022multihiertt} reasons from heterogeneous information and is more challenging as it often requires numerical reasoning to answer the question in addition to span extraction.

\begin{table}[t]
\small
  \centering
    \begin{tabular}{lcc}
    \toprule
     & EM    & F1 \\
    \midrule
    1 step & 43.62 & 47.80 \\
    2 steps & 34.67 & 37.91 \\
    3 steps & 22.43 & 24.57 \\
    >3 steps & 15.14 & 17.19 \\
    \midrule
    Full Results & 36.22 & 38.43 \\
    \bottomrule
    \end{tabular}%
  \caption{Results of different numerical reasoning steps using MT2Net (RoBERTa-large) on the test set of MultiHiertt.}
  \label{tab:steps}%
\end{table}%

To empower hybrid tabular-textual QA models with numerical reasoning ability, TAGOP \citep{zhu2021tat} uses sequence tagging to extract supporting facts, then performs a single arithmetic operation with one of a number of pre-defined operators. To support multi-step reasoning, the encoder-decoder transformer such as T5 \citep{raffel2020exploring} or decoder transformer such as GPT-2 \citep{radford2019language} can be used to autoregressively decode program sequences, but previous work \citep{chen2022convfinqa} has verified that using pre-trained models does not have advantages in numerical reasoning. FinQANet \citep{chen2021finqa} and MT2Net \citep{zhao2022multihiertt} use the RoBERTa model \citep{liu2019roberta} as the encoder and a specially designed LSTM decoder with structural preservation of the program autoregressively decode program sequences, and each decoding step produces either an operator or an operand, achieving the current state-of-the-art performance on ConvFinQA \citep{chen2022convfinqa} and MultiHiertt \citep{zhao2022multihiertt} datasets, respectively. However, this step-by-step autoregressive decoding process suffers from severe exposure bias. During training, the model uses gold references as decoding history (teacher forcing), and learns to rely on the reference decoding history. But the decoding history is very likely to be wrong during inference if the model cannot produce high quality predictions, and wrong predictions in early steps may negatively affect subsequent predictions and lead to further errors in following steps \citep{zhang-etal-2019-bridging}. Unfortunately, this is the case with hybrid QA, where the prediction performance of current methods are far from good, and as a result of error propagation, program generation accuracy drops heavily as the number of decoding steps increases (Table~\ref{tab:steps}).

Non-autoregressive generation can relieve the dependency on prediction results of earlier steps, addressing the exposure bias issue, while boosting the generation speed with better parallelization. In this paper, we propose a \textbf{N}on-\textbf{A}utoregressive \textbf{P}rogram \textbf{G}eneration model (NAPG). Instead on generating the program sequence in a step-by-step manner, we only use the encoder representation (without decoding history), and employ an independent numerical reasoning tuple (operator, operands) generator for each reasoning step to predict the operator and its operands. The numerical reasoning tuple generator contains a soft masking mechanism \citep{zhang2020spelling} to derive its specific input representation from the encoder representation by highlighting the operand representations of the step, followed by an operator generator, an operand generator and an order predictor. We also utilize a length predictor to control the number of numerical reasoning tuples produced. As the numerical reasoning tuple generator does not leverage any previous decoder history steps, our method prevents the generation with the exposure bias problem and greatly improves the generation speed due to parallelization.

Our main contributions are as follows:

\begin{itemize}
    \item We propose a non-autoregressive program generation model (NAPG), which can generate the full reasoning programs in parallel. Compared to previous autoregressive generators, our method does not suffer from the exposure bias issue and is much faster due to parallelization.
    \item In our experiments on the ConvFinQA \citep{chen2022convfinqa} and MultiHiertt \citep{zhao2022multihiertt} datasets, the NAPG model can bring about substantial improvements over the strong FinQANet ($+5.06$ Exe Acc and $+4.80$ Prog Acc points) and MT2Net ($+7.97$ EM and $+6.38$ F1 points) baselines, establishing the new state-of-the-art performance, while being $\sim$21 times as fast in program generation. Our further analysis shows that, the performance loss of NAPG is also significantly smaller than the baseline with increasing numbers of numerical reasoning steps.
\end{itemize}

\begin{figure}[t]
    \centering
    \includegraphics[width=1.0\linewidth]{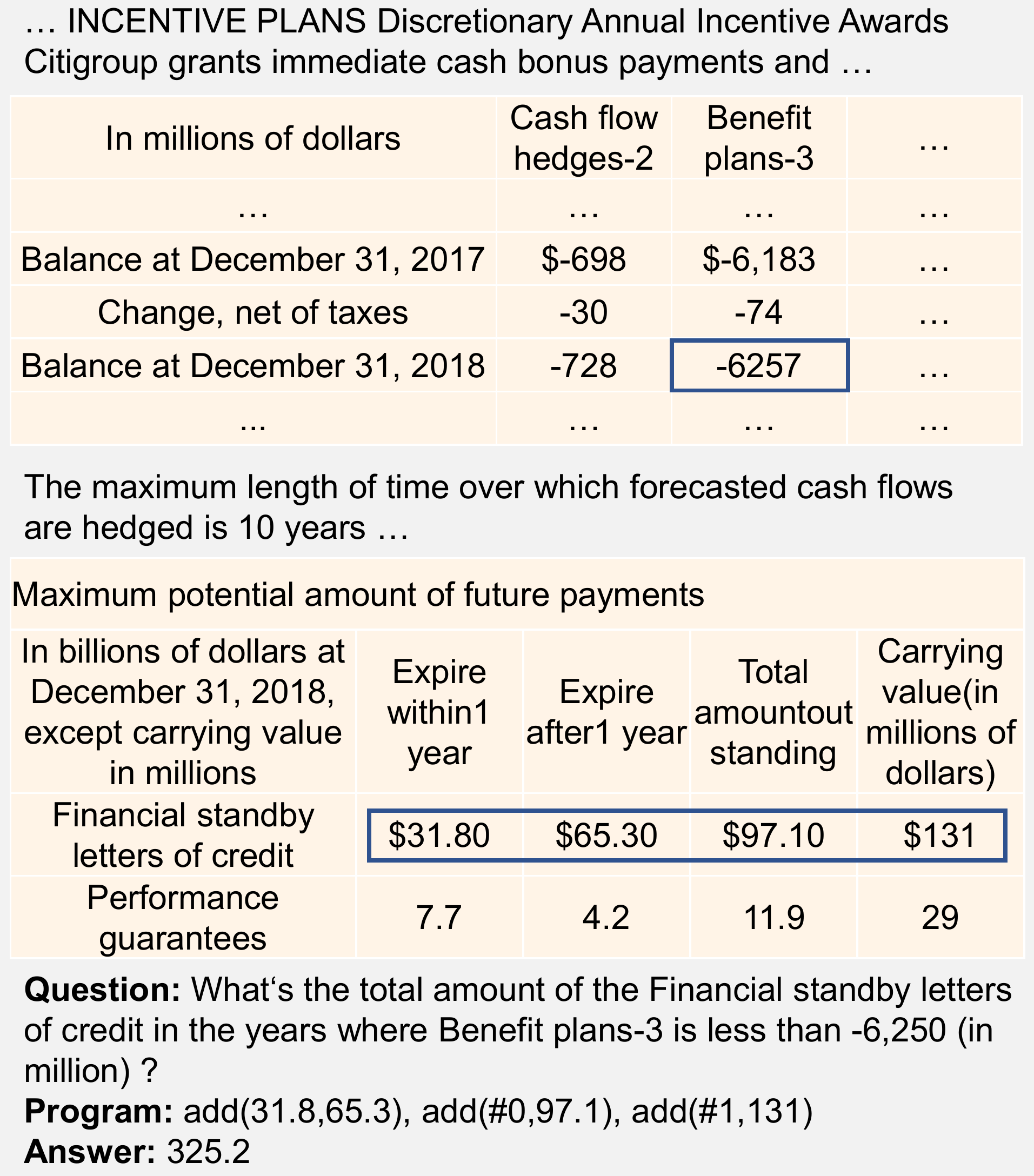}
    \caption{An example from the MultiHiertt dataset. In the numerical reasoning question, the system needs to locate which year has less than -6,250 Benefit plans-3 from the first table, and then select the relevant numbers from the second hierarchical table as operands to calculate the answer with addition as the operator. Better viewed in color, the supporting facts are in light blue boxes.}
    \label{fig:1}
\end{figure}
\begin{figure*}[t]
    \centering
    \includegraphics[width=0.9\linewidth]{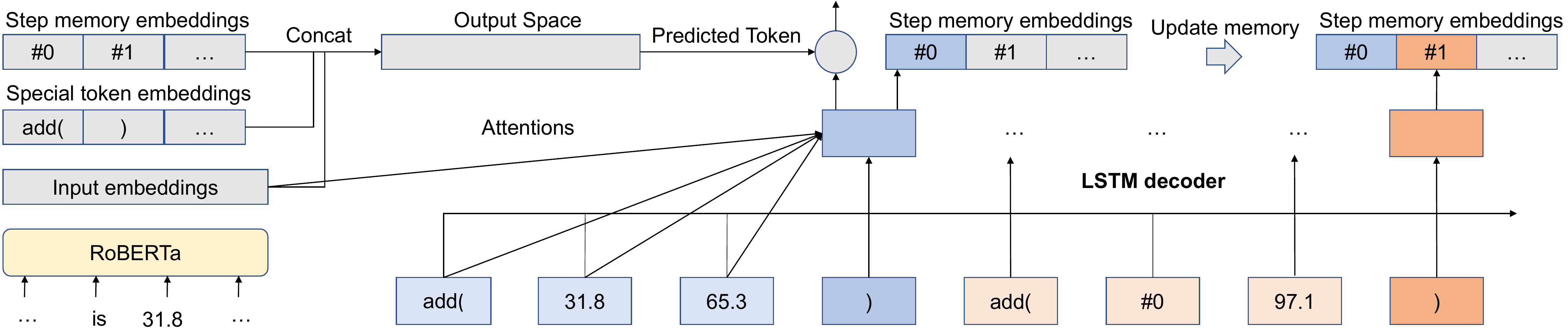}
    \caption{Autoregressive Program Generation for Numerical Reasoning.}
    \label{fig:2}
\end{figure*}

\section{Preliminaries}

\paragraph{Task Description} Question answering over hybrid tabular textual data requires reasoning from heterogeneous information, involving numerical reasoning or span extraction. As shown in Figure~\ref{fig:1}, given the question $Q$, the system is to find its answer from tables $T$ and texts $E$. For some cases, the model only needs to extract an answer span $A$ from the input. For many other cases involving numerical reasoning, the model has to generate a program sequence $G = \left\{ {{g_0},{g_1},...,{g_n}} \right\}$, where ${g_i}$ stands for the token of the program, which is either extracted from the input, or selected from pre-defined special tokens, including operators and special operands, and the probability of an answer $A$ is calculated by summing over the probabilities of all programs $G_i$ from which the answer $A$ can be obtained:

\begin{equation}
P\left( {A|T,E,Q} \right) = \sum\limits_i {P\left( {{G_i}|T,E,Q} \right)} 
\label{eqa:task}
\end{equation}

\paragraph{Fact Retrieving}  MT2Net converts data cells of tables into sentences with their row and column headers. Due to the input length limitation of PLMs, MT2Net first concatenates the question with each sentence as input to train a BERT-based binary-classifier (bi-classifier) for supporting fact classification. Next, it takes the top $n$ sentences based on the supporting fact classification prediction as the input for the next stage. Another classifier is used to determine whether the next stage is span extraction or numerical reasoning.

\paragraph{Span Extraction} MT2Net uses the T5-base model \citep{raffel2020exploring} for span extraction questions, where the model takes the concatenation of the question and the sentences containing supporting facts as input, and generates the answer spans.

\paragraph{Autoregressive Numerical Reasoning} MT2Net first uses RoBERTa as an encoder to obtain the context-aware representations of the question and the sentences containing supporting facts, and concatenates them with the embeddings of pre-defined special tokens, such as the function names, predefined constants, etc. Next, it uses an LSTM decoder to generate the program sequence for the deduction of the answer. Each decoding step makes predictions over the concatenated matrix and selects either an operator or an operand, as shown in Figure~\ref{fig:2}. FinQANet uses the same module for autoregressive numerical reasoning like MT2Net, but it does not use an additional classifier to predict question type during fact retrieving, nor does it have span extraction module.

\begin{figure*}[t]
    \centering
    \includegraphics[width=\linewidth]{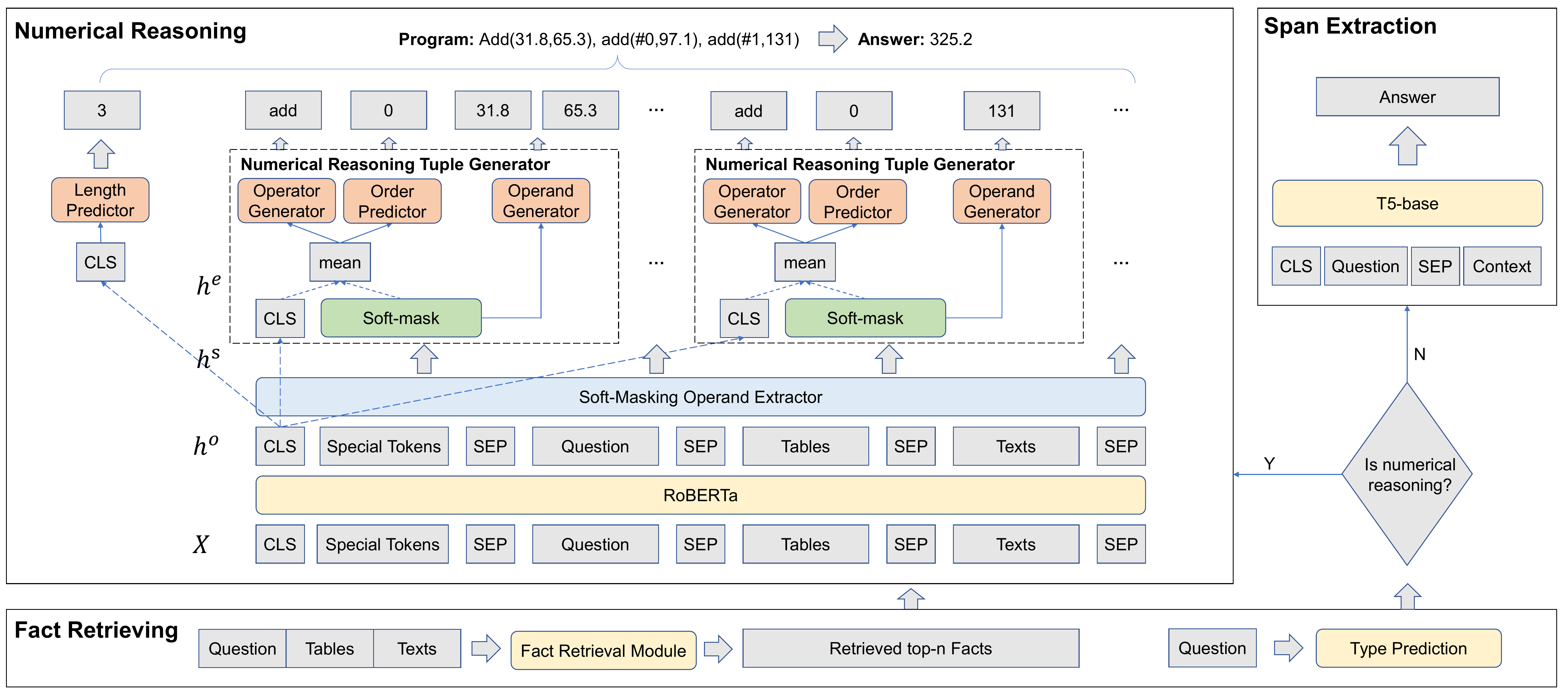}
    \caption{The NAPG Model. For multi-step reasoning, the Soft-Masking Operand Generator is only computed once and extracts all operands of all reasoning steps, there are multiple different Numerical Reasoning Tuple Generators and each Numerical Reasoning Tuple Generator independently picks its operands, operator, order of each reasoning step.}
    \label{fig:3}
\end{figure*}

\section{Our Approach}

We present our non-autoregressive program generation model, which can generate the full program sequences independently to address the exposure bias issue of the step-by-step autoregressively program generation model, and speed up the generation by supporting better parallelization. The NAPG model framework is shown in Figure~\ref{fig:3}. It first uses a bi-classifier to retrieve the most relevant facts, and uses another bi-classifier to identify the question type like MT2Net. For span extraction questions, NAPG uses the same T5-base model as MT2Net to generate the answer given the concatenation of the question and the sentences containing supporting facts. But for numerical reasoning, we design a non-autoregressive approach to program generation, which is quite different from the autoregressive LSTM decoder used by MT2Net. Especially, NAPG can also use numerical reasoning modules to solve span extraction questions without the additional question classification and span extraction models.

\subsection{Non-Autoregressive Program Generation}

Our non-autoregressive program generation is implemented by a length predictor to predict the number of numerical reasoning steps (number of program tuples), a soft-masking operand extractor to identify all operands in the program sequence, and a set of modules for the program generation of all steps, where each reasoning step includes a soft-masking operand generator to select the operands for the operator, an operator generator to predict the operator, and an order predictor to decide the order of the operands. As the program tuples (each containing one operator, two operands and the order of the two operands) of different programming steps are likely to be different, we use the same architectures but independent modules for different steps.

Since operations such as averaging and numerical order conversion may occur during reasoning, we add constants within 10 and common order values into special tokens following FinQANet and MT2Net, and concatenate special tokens with the question and the sentences containing supporting facts as input to the RoBERTa encoder.

\paragraph{Length Predictor} Numerical reasoning requires a variable number of inference steps, so we employ a multi-class classifier as the length predictor to predict the number of reasoning steps, where each reasoning step contains a complete program tuple (operator, operands, order). The classifier is an FFN layer with the RoBERTa representation of the $[\text{CLS}]$ token without soft masking as input.

\begin{equation}
{p^{\text{length}}} = \text{softmax}\left( {\text{FFN}\left( {\left[ {\text{CLS}} \right]} \right)} \right)
\end{equation}
\noindent where $\text{FFN}$ is a 2-layer feed-forward network with GELU \citep{hendrycks2016bridging} as the activation function.

\paragraph{Soft-Masking Operand Extractor} Extracting correct operands is crucial for the inference of the correct answer. We empirically find that extracting \textbf{all operands} for the program sequence prior to the Numerical Reasoning Tuple Generator benefits the performance (\$~\ref{sec:appendrs}), and we employ an FFN as the Soft-Masking Operand Extractor layer over the full RoBERTa representation to extracting all operands.

\begin{equation}
{p^t} = \text{softmax}\left( {\text{FFN}\left( {{\bm{h}^o}} \right)} \right)
\end{equation}

\noindent where $\text{FFN}$ is a 2-layer feed-forward network with GELU as the activation function. $\bm{h}^o$ is the RoBERTa representation. $p^t$ represents the probability that the token is an operand.

Then we soft mask \citep{zhang2020spelling} the RoBERTa representation with $p^t$.

\begin{equation}
{\bm{h}^s} = {\bm{h}^o}*p^t + {\bm{v}^m}*(1 - p^t)
\label{eqa:mask}
\end{equation}

\noindent where $\bm{h}^s$ is the soft-masked representation, $\bm{v}^m$ stands for the mask embedding, and ``*'' indicates element-wise multiplication.

A large $p^t$ would make the soft masking result close to the original embedding, while a small $p^t$ would turn the result close to the mask embedding. The soft-masking mechanism can thus represent the operands with a higher priority. Compared to using the classification results for hard masking, soft masking is differentiable and can be trained in an end-to-end manner while alleviating the error propagation issue. We use a zero vector with all dimensions set to 0 as the mask embedding, as we empirically find that this works better than the other options (including the embedding of the special [MASK] token) for this task.

Especially, when answering span extraction questions, we predict that the length is 0, and then generate the answer spans from the outputs of the soft masking operand extractor.

\paragraph{Operand Generator} We also utilize the soft masking mechanism to extract the \textbf{two operands} of the specific reasoning step from the input in the Numercical Reasoning Tuple Generator. Compared to the soft-masking Operand Extractor that identifies all operands of the complete program sequences, the Operand Generator only finds the two operands of the specific program step by selecting only two tokens with the highest prediction probabilities from either the numbers in the retrieved sentences or the set of pre-specified tokens representing the computation results of preceding program steps or pre-defined numbers.

\begin{equation}
p^e = \text{softmax}\left( {\text{FFN}\left( {{\bm{h}^s}} \right)} \right)
\end{equation}

\begin{equation}
\bm{h}^e = {\bm{h}^s}*p^e + \bm{v}^m*\left( {1 - p^e} \right)
\end{equation}

\noindent where $\bm{h}^s$ and $\bm{h}^e$ are the soft-masked representation from the Operand Extractor and the soft masking result respectively. $p^e$ represents the probability that the token is the operand of the step.

\paragraph{Operator Generator} We define six operators following FinQANet and MT2Net: Addition, Subtraction, Multiplication, Division, Exp, Greater. We average the soft-masked representations produced by the operand generator of the reasoning step and the embedding of the $[\text{CLS}]$ token as input, and use a multi-classifier as the operator generator to select the operator.

\begin{equation}
{p^{\text{op}}} = \text{softmax}\left( {\text{FFN}\left( {\text{mean}\left( {{\left[ {\text{CLS}} \right]|\bm{h}^e}} \right)} \right)} \right)
\end{equation}

\paragraph{Order Predictor} The order of the two operands matters when the operator is subtraction, division, exp or greater. We also take the mean pooling of the $[\text{CLS}]$ token embedding and the soft-masked embeddings produced by the operand generator of the reasoning step as input, and use a bi-classifier to predict the order of the operands (whether their order is as in the input or in the reverse order).

\begin{equation}
p^{\text{order}} = \text{softmax}\left( {\text{FFN}\left( {\text{mean}\left( { {\left[ {\text{CLS}} \right]|\bm{h}^e} } \right)} \right)} \right)
\end{equation}

\subsection{Training}

To jointly optimize all objectives for numerical reasoning, we minimize the weighted sum of the negative log-likelihood losses of individual modules.

\begin{equation}
\small
\begin{array}{l}
\mathcal{L}{\rm{ = }}{\lambda ^t}*\text{NLL}\left( {\log \left( {{p^t}} \right),{r^t}} \right)\\
 + {\lambda ^{\text{length}}}*\text{NLL}\left( {\log \left( {{p^{\text{length}}}} \right),{r^{\text{length}}}} \right)\\
 + {\lambda ^e}*\sum\limits_{i = 0}^n {\text{NLL}\left( {\log \left( {p_i^e} \right),r_i^e} \right)} \\
 + {\lambda ^{\text{op}}}*\sum\limits_{i = 0}^n {\text{NLL}\left( {\log \left( {p_i^{\text{op}}} \right),r_i^{\text{op}}} \right)} \\
 + {\lambda ^{\text{order}}}*\sum\limits_{i = 0}^n {\text{NLL}\left( {\log \left( {p_i^{\text{order}}} \right),r_i^{\text{order}}} \right)} 
\end{array}
\label{eqa:tloss}
\end{equation}

\noindent where $\text{NLL}$ stands for the negative log-likelihood loss function, $r$ indicates the ground truths, $\lambda$ represents the weight of each module, and $n$ is the maximum number of reasoning steps.

\subsection{Discussions}

The general design of NAPG only involves element-wise computations, the single FFN for length prediction, and 3 sets of FFNs with different parameters but the same architecture for operand generation, operation generation and order prediction respectively. When generating the program tuple sequence, the length predictor only needs to be computed once, and the element-wise computations can be easily parallelized. As for each set of FFNs with different parameters but the same architecture, their activation function can be easily parallelized, and their linear layers with different parameters can be parallelized with the batch matrix-matrix multiplication function implemented in almost all modern linear algebra libraries.

Compared to autoregressive decoding, non-autoregressive counterparts ignore the decoding history which may have a potential negative impact on its coherence, but in case of program generation for numerical reasoning, coherence may be affected less, while \textbf{the autoregressive decoding is very likely to be mislead especially for program generation given the prediction quality is not high ($<50\%$, as shown in Figure~\ref{fig:4}).}

\section{Experiments}

\subsection{Settings}

\textbf{Dataset} We conducted our experiments on the ConvFinQA \citep{chen2022convfinqa} and MultiHiertt \citep{zhao2022multihiertt} datasets. The ConvFinQA dataset poses a great challenge in modeling long-range, complex numerical reasoning paths in real-world conversations \citep{chen2022convfinqa}. Compared with existing datasets, each document in MultiHiertt contains multiple hierarchical tables and longer unstructured text. A more complex reasoning process across multiple tables and paragraphs is required to correctly answer questions \citep{zhao2022multihiertt}. The test set labels of the ConvFinQA and MultiHiertt datasets are not public. Table~\ref{tab:datasets} shows that the ConvFinQA dataset poses great challenge in modeling long-range, complex numerical reasoning paths in real-world conversations. Compared to ConvFinQA, each document in MultiHiertt contains more hierarchical tables and longer unstructured text, so it is more difficult to correctly answer the question.

\begin{table*}[htbp]
\small
  \centering
    \begin{tabular}{lccccccc}
    \toprule
    Dataset & Avg. \# text words & Table types  & Avg. \# tables & Turn  & \# Train & \# Dev & \# Test \\
    \midrule
    ConvFinQA & 675.61  & Flat  & 1     & Multi & 11,104 & 1,490 & 1,521 \\
    MultiHiertt & 1,645.9 & Hierarchical & 3.89  & Single & 7,830 & 1,044 & 1,566 \\
    \bottomrule
    \end{tabular}%
  \caption{Statistics of ConvFinQA and MultiHiertt datasets.}
  \label{tab:datasets}%
\end{table*}%

\paragraph{Evaluation Metrics} We evaluated the performance by Exact Matching (EM) and the adopted numeracy-focused F1 \citep{dua2019drop} for MultiHiertt, and execution accuracy (Exe Acc) and program accuracy (Prog Acc) for ConvFinQA following previous work. Exe Acc calculates the accuracy of the execution results of the prediction program. Prog Acc calculates the accuracy of the prediction program (both operators and operands) with the ground program, which is stricter because there may be multiple different programs that can execution the same results.

\paragraph{Baselines} GPT-2 \citep{radford2019language} and T5 \citep{raffel2020exploring} are two generative models. TAGOP \citep{zhu2021tat} first uses the sequence tagging method to extract facts, then performs only one arithmetic operation with pre-defined operators. FinQANet \citep{chen2021finqa} and MT2Net \citep{zhao2022multihiertt} are able to perform multi-step reasoning, and they both use an autoregressive LSTM decoder to generate the program.

\paragraph{Model Settings} We tuned hyper-parameters on the development set ($\S$~\ref{ssec:ashp}). To fairly compare with existing state-of-the-art results, GPT-2 and T5 use medium and large respectively, the rest of baselines are based on the RoBERTa-large model. To focus on program generation and for fair comparison, we only replace the program generation module of MT2Net with NAPG and leave the other parts unchanged. As FinQANet uses a single LSTM to decode either the program sequence or the span to extract according to question type without having an individual span extraction module, we ask the length predictor of NAPG to predict a length of 0 and extract the span with the highest prediction probability directly from the output of the operand extractor in this case to take care of span extraction questions on the ConvFinQA dataset. To fairly compare with existing state-of-the-art results, we used the same hyper-parameters as the SotA model. We train NAPG (RoBERTa-large) with batch size of 6 on a single RTX3090 GPU on both ConvFinQA and MultiHiertt datasets. We set the maximum number of reasoning steps n to 5, 10 for ConvFinQA and MultiHiertt respectively. It takes about 12 hours to train the numerical reasoning model of NAPG (RoBERTa-large) on MultiHiertt, and it takes about 10 hours to train the numerical reasoning model of NAPG (RoBERTa-large) on ConvFinQA. For base model, $\lambda ^{\text{op}}$ was set to 2, $\lambda ^{\text{order}}$ was set to 1.5, $\lambda ^{\text{t}}$ was set to 1.2, $\lambda ^{\text{length}}$ was set to 1.1, and $\lambda ^{\text{e}}$ was set to 1 for ConvFinQA, $\lambda ^{\text{op}}$ was set to 2, the others were all set to 1 for MultiHiertt. For large model, $\lambda ^{\text{op}}$ was set to 2, and $\lambda ^{\text{order}}$ was set to 1.5, the others were set to 1 for both ConvFinQA and MultiHiertt datasets.

\begin{table}[t]
\small
  \centering
    \begin{tabular}{lcc}
    \toprule
          & Exe Acc    & Prog Acc \\
    \midrule
    GPT-2 (medium) & 58.19 & 57.00 \\
    T5 (large) & 58.66 & 57.05\\
    FinQANet (RoBERTa-large) & 68.90 & 68.24 \\
    \midrule
    Ours (RoBERTa-base) & 69.82 & 68.84 \\
    Ours (RoBERTa-large) & \textbf{73.96} & \textbf{73.04} \\
    \bottomrule
    \end{tabular}%
  \caption{Main results on ConvFinQA.}
  \label{tab:conv}%
\end{table}%

\begin{table}[t]
\small
  \centering
    \begin{tabular}{lcc}
    \toprule
          & EM    & F1 \\
    \midrule
    TAGOP (RoBERTa-large) & 17.81 & 19.35 \\
    FinQANet (RoBERTa-large) & 31.72 & 33.60 \\
    MT2Net (RoBERTa-large) & 36.22 & 38.43 \\
    \midrule
    Ours (RoBERTa-base) & 38.19 & 38.81 \\
    Ours (RoBERTa-large) & \textbf{44.19} & \textbf{44.81} \\
    \bottomrule
    \end{tabular}%
    \caption{Main results on MultiHiertt.}
  \label{tab:multi}%
\end{table}%

\begin{table}[t]
\small
  \centering
    \begin{tabular}{lcc}
    \toprule
          & \multicolumn{2}{c}{Dev} \\
          & EM    & F1 \\
    \midrule
    MT2Net (RoBERTa-large) & 41.35 & 41.35 \\
    Ours (RoBERTa-large) & \textbf{48.20} & \textbf{48.20} \\
    \bottomrule
    \end{tabular}%
  \caption{Main results of numerical reasoning on MultiHiertt.}
  \label{tab:num}%
\end{table}%

\subsection{Main Results}

We first compare NAPG (with both base and large settings) with our baselines. Results are shown in Tables~\ref{tab:conv} and \ref{tab:multi}.

Tables~\ref{tab:conv} and \ref{tab:multi} show that: 1) using pre-trained models for numerical reasoning program generation does not lead to better performance than LSTM. 2) already with the RoBERTa base setting, our method is able to achieve better performances on both datasets. 3) using the RoBERTa large setting can further boost the performance of NAPG, and lead to large improvements on both ConvFinQA ($+5.06$/$+4.80$ Exe/Prog Acc points) and MultiHiertt ($+7.97$/$+6.38$ EM/F1 points) datasets.

As our approach only modifies the program generation part, we also tested the performance of NAPG and MT2Net on all numerical reasoning questions in the development set of MultiHiertt. Results are shown in Table~\ref{tab:num}.

Table~\ref{tab:num} shows that NAPG can lead to large improvements over the MT2Net baseline ($+6.85$ EM and F1) in numerical reasoning.
\subsection{Performance w.r.t. Reasoning Steps}

\begin{figure*}[t]
    \centering
    \includegraphics[width=\linewidth]{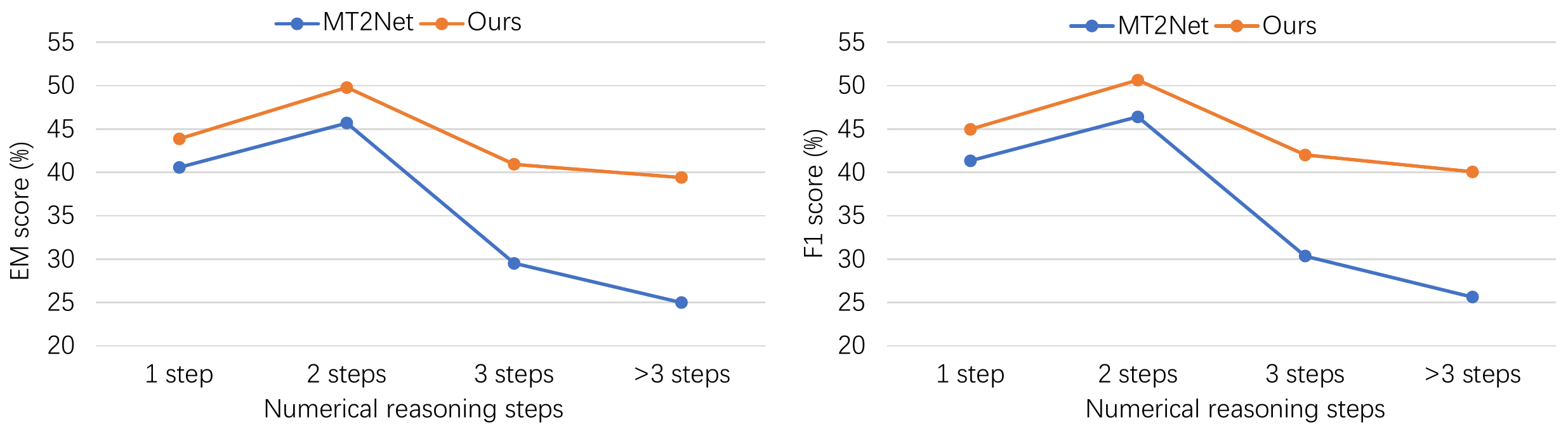}
    \caption{Performance of different numerical reasoning steps on the development set of MultiHiertt.}
    \label{fig:4}
\end{figure*}

To verify whether NAPG can really address the error accumulation issue of autoregressive program generation, we analyze the performance of NAPG and MT2Net w.r.t. different numbers of reasoning steps. For fairness, we used RoBERTa-large as the encoder of both NAPG and MT2Net. As the test set is not publicly available, our analysis is performed on the development set of MultiHiertt and the results of MT2Net are from \citet{zhao2022multihiertt}. Results are shown in Figure~\ref{fig:4}.

Figure~\ref{fig:4} shows that: 1) despite the metrics reporting highest scores with 2 reasoning steps, the general performance trend is descending while increasing the number of reasoning steps. 2) our NAPG approach outperforms the MT2Net in all aspects by a large margin. 3) as the number of reasoning steps increases, the improvements of our method are much larger over the autoregressive MT2Net baseline ($+11.41$/$+11.67$ EM/F1 when the number of reasoning steps is 3 and $+14.42$/$+14.43$ EM/F1 when it is larger than 3).

The performance drop with increased numbers of reasoning steps with our non-autoregressive method is much smaller than the autoregressive MT2Net, showing the advantage of NAPG in handling questions that require inference with long program sequences. Intuitively, in the generation of longer program sequences, the autoregressive model is more likely to suffer from exposure bias, while the non-autoregressive generation prevents our method from suffering from this issue.

\subsection{Performance of NAPG without Span Extraction Model}
\label{sec:napg}

\begin{table}[t]
\small
  \centering
    \begin{tabular}{lcc}
    \toprule
          & {Dev} & {Test}\\
          & EM/F1  & EM/F1 \\
    \midrule
    MT2Net (RoBERTa-base) & 35.69/37.81 & 34.32/36.17 \\
    NAPG (RoBERTa-base) & 39.27/40.21 & 38.19/38.81 \\
    NAPG$^*$ (RoBERTa-base) & \textbf{41.37/42.24} & \textbf{38.44/39.58} \\
    \midrule
    MT2Net (RoBERTa-large) & 37.05/39.96 & 36.22/38.43 \\
    NAPG (RoBERTa-large) & 45.79/46.15 & 44.19/44.81 \\
    NAPG$^*$ (RoBERTa-large) & \textbf{45.88/47.27} & \textbf{45.59/47.04} \\
    \bottomrule
    \end{tabular}%
    \caption{Results of NAPG and NAPG$^*$ on MultiHiertt.}
  \label{tab:napg*}%
\end{table}%
To focus on program generation and for fair comparison, we only replace the program generation module of MT2Net with NAPG and leave the other parts unchanged. But NAPG can also use numerical reasoning modules to solve span extraction questions without the additional question classification and span extraction models (we call it as NAPG$^*$). We tested the performance of NAPG and NAPG$^*$ using the same parameters on the MultiHiertt dataset. Results are shown in Table~\ref{tab:napg*}.

Table~\ref{tab:napg*} shows that NAPG$^*$ is able to achieve better performances on both base and large settings, perhaps because there are fewer span extraction questions in the MultiHiertt dataset (1524/7830 on the train set), using the two types of problems to train NAPG$^*$ model together will enable the model to learn more knowledge.

\subsection{Performance of Soft Masking Operand Extractor}
\label{sec:appendrs}

\begin{table}[t]
\small
  \centering
    \begin{tabular}{lcc}
    \toprule
          & EM    & F1 \\
    \midrule
    MT2Net & 36.22 & 38.43 \\
   \ \  + Soft Masking Operand Extractor & 41.51 & 41.88 \\
    \midrule
    NAPG  & \textbf{44.19} & \textbf{44.81} \\
    \bottomrule
    \end{tabular}%
  \caption{Performance of Soft Masking Operand Extractor on Multihiertt.}
  \label{tab:appendrs}%
\end{table}%

To test the effectiveness of the soft masking operand extraction and non-autoregressive decoding separately, we also add the Soft-Masking Operand Extraction module into the original MT2Net with the RoBERTa-large setting. Results are shown in Table~\ref{tab:appendrs}.

Table~\ref{tab:appendrs} shows that the Soft-Masking Operand Extraction module can benefit the performance of MT2Net, and our non-autoregressive decoding method can lead to further improvements.

\begin{table}[t]
\small
  \centering
    \begin{tabular}{l|cc}
    \toprule
          & MT2Net & Ours \\
          & EM/F1 & EM/F1 \\
    \midrule
    text-only questions & 34.78/34.78 & \textbf{46.96/46.96} \\
    table-only questions & 40.83/41.95 & \textbf{46.75/47.98} \\
    \ \ \ \  with $\geq$ 2 tables & 75.86/75.86 & \textbf{86.21/86.21} \\
    table-text questions & 40.27/40.94 & \textbf{45.01/45.96} \\
    \ \ \ \  with $\geq$ 2 tables & 71.70/71.70 & \textbf{72.64/72.64} \\
    \midrule
    Full Results & 39.85/40.59 & \textbf{45.79/46.72} \\
    \bottomrule
    \end{tabular}%
    \caption{Performance of different question types on the development set of MultiHiertt.}
  \label{tab:abl}%
\end{table}%

\subsection{Performance Analysis w.r.t. Question Type}

We tested the ability of NAPG and MT2Net in handling different sources of supporting facts with the RoBERTa-large setting. Results are shown in Table~\ref{tab:abl}.

Table~\ref{tab:abl} shows that NAPG outperforms the strong MT2Net baseline in all evaluations. Text-only questions are not span extraction but all of them are   numerical reasoning questions in the MultiHiertt dataset, our NAPG method greatly improves the performance of numerical reasoning and leads to such substantial improvements (+$12.18$ EM and F1). Given that table-only questions normally require numerical reasoning, and that the number of reasoning steps positively co-relates with the number of tables, the large improvements (+$10.35$ EM and F1) over the strong baseline ($75.86$ EM and F1) for table-only questions with no less than 2 tables confirm the advantage of NAPG in the numerical reasoning of complex questions.

\begin{table}[t]
\small
  \centering
    \begin{tabular}{ccccc|cc}
    \toprule
          &       &       &       &       & \multicolumn{2}{c}{Dev} \\
          &       &       &       &       & base  & large \\
    $\lambda ^\text{t}$    &$\lambda ^{\text{l}}$ & $\lambda ^\text{e}$    & $\lambda ^{\text{op}}$   & $\lambda ^{\text{od}}$ & EM/F1 & EM/F1 \\
    \midrule
    1     & 1     & 1     & 1     & 1     & 38.60/39.54 & 44.35/45.29 \\
    \midrule
    2     &       &       &       &       & 37.84/38.77 & \textbf{44.92/45.86} \\
          & 2     &       &       &       & 37.45/38.39 & \textbf{44.64/45.57} \\
          &       & 2     &       &       & 37.93/38.87 & 42.72/43.66 \\
          &       &       & 2     &       & \textbf{39.27/40.21} & 43.77/44.71 \\
          &       &       &       & 2     & \textbf{39.18/40.11} & 43.87/44.81 \\
    \midrule
    2     & 1.5   &       &       &       & 37.55/38.49 & 45.21/46.15 \\
          &       &       & 2     & 1.5   & \textbf{37.93/38.87} & \textbf{45.79/46.72} \\
    \bottomrule
    \end{tabular}%
    \caption{Effects of different weights of each module. $\lambda ^{\text{l}}$ and $\lambda ^{\text{od}}$ stand for $\lambda ^{\text{length}}$ and $\lambda ^{\text{order}}$ respectively in Eq.~\ref{eqa:tloss}.}
  \label{tab:lamda}%
\end{table}%

\subsection{Ablation Study of Hyper-Parameters}
\label{ssec:ashp}

To study the effects of different components in NAPG on the performance, we explored a number of hyper-parameter values for the combination of training losses (Eq.~\ref{eqa:tloss}) on the development set of MultiHiertt. Specifically, we first increase only one of all hyper-parameters to 2 while keeping the others set to 1 in each experiment, and then increase all hyper-parameters that lead to improvements for either the base setting or the large setting together, while assigning the hyper-parameter that leads to more improvements a larger value. Results are shown in Table~\ref{tab:lamda}.

Table~\ref{tab:lamda} shows that the best performing settings are different with different model settings. The best setting among all tested cases for the base setting is to use a $\lambda ^{\text{op}}$ of 2 while setting the others to 1, and for the large setting is to use a $\lambda ^{\text{op}}$ of 2, a $\lambda ^{\text{order}}$ of 1.5 while setting the others to 1.

\begin{table}[t]
\small
  \centering
    \begin{tabular}{ccc}
    \toprule
    Model & Time (s) & Speed-up \\
    \midrule
    LSTM  & 168.86 & 1x \\
    Ours  & 8.04  & 21x \\
    \bottomrule
    \end{tabular}%
  \caption{Time costs for program generation.}
  \label{tab:time}%
\end{table}%

\subsection{Program Generation Speed Analysis}

Non-autoregressive program generation allows our approach to benefit from parallelization. We compared the program generation speed of NAPG and the LSTM decoder of MT2Net by recording the time costs of the program generation modules on all numerical reasoning questions in the training set of MultiHiertt. Results are shown in Table~\ref{tab:time}.

Table~\ref{tab:time} shows that NAPG is 21 times as fast as MT2Net, showing the substantial advantage of non-autoregressive decoding over autoregressive decoding in terms of speed due to parallelization. And we also test the RoBERTa consumes 67.95 seconds for one epoch on the training set.

\subsection{Case Study}

\begin{table*}[htbp]
\small
  \centering
    \begin{tabular}{c|p{33.815em}}
    \toprule
    \multicolumn{1}{c|}{\multirow{4}[2]{*}{Type \uppercase\expandafter{\romannumeral1}}} & \textbf{Question:} How much did the company 2019s valuation allowance decrease from 2011 to 2012? \\
          & \textbf{Reference:} -0.09542 subtract(19520,21579), divide(\#0,21579) \\
          & \textbf{MT2Net:} -0.10548 ['subtract(', '19520.0', '21579.0', ')', 'divide(', '\#0', '19520.0', ')', 'EOF'] \\
          & \textbf{Ours:} -0.09542 ['subtract(', '19520.0', '21579.0', ')', 'divide(', '\#0', '21579.0', ')'] \\
    \midrule
    \multicolumn{1}{c|}{\multirow{4}[2]{*}{Type\uppercase\expandafter{\romannumeral2}}} & \textbf{Question:} What's the total amount of the U.S. dollars sold for Pounds sterling in the years where U.S. dollars sold for Pounds sterling is greater than 1? \\
          & \textbf{Reference:} 735.0 add(390,268), add(\#0,77) \\
          & \textbf{MT2Net:} 658.0 ['add(', '390.0', '268.0', ')', 'EOF'] \\
          & \textbf{Ours:} 735.0 ['add(', '390.0', '268.0', ')', 'add(', '\#0', '77.0', ')'] \\
    \midrule
    \multicolumn{1}{c|}{\multirow{4}[2]{*}{Type \uppercase\expandafter{\romannumeral3}}} & \textbf{Question:} How much of profit before taxes is there in total (in 2017) without U.S. tax reform impact and Gain on sale of equity investment? (in million) \\
          & \textbf{Reference:} 5639.0 add(4082,1256), add(\#0,301) \\
          & \textbf{MT2Net:} 5554.0 ['add(', '4082.0', '1256.0', ')', 'add(', '\#0', '301.0', ')', 'subtract(', '\#1', '85.0', ')', 'EOF'] \\
          & \textbf{Ours:} 5639.0 ['add(', '4082.0', '1256.0', ')', 'add(', '\#0', '301.0', ')'] \\
    \midrule
    \multicolumn{1}{c|}{\multirow{4}[2]{*}{Type \uppercase\expandafter{\romannumeral4}}} & \textbf{Question:} What's the average of the Fuel for Amount in the years where Wheelabrator is positive? \\
          & \textbf{Reference:} 626.66667 add(603,649), add(\#0,628), divide(\#1,3) \\
          & \textbf{MT2Net:} 626.0 ['add(', '603.0', '649.0', ')', 'divide(', '\#0', 'const\_2', ')', 'EOF'] \\
          & \textbf{Ours:} 626.66667 ['add(', '603.0', '649.0', ')', 'add(', '\#0', '628.0', ')', 'divide(', '\#1', '3.0', ')'] \\
    \midrule
    \multicolumn{1}{c|}{\multirow{4}[2]{*}{Type \uppercase\expandafter{\romannumeral5}}} & \textbf{Question:} What's the total amount of U.S. large cap stocks , U.S. small cap stocks, Non-U.S. large cap stocks and Non-U.S. small cap stocks in 2013? (in million) \\
          & \textbf{Reference:} 273.0 add(140,56), add(\#0,56), add(\#1,21) \\
          & \textbf{MT2Net:} 322.0 ['add(', '140.0', '56.0', ')', 'add(', '\#0', '21.0', ')', 'add(', '\#1', '21.0', ')', 'add(', '\#2', '21.0', ')', 'add(', '\#3', '21.0', \\
          & \textbf{Ours:} 273.0 ['add(', '140.0', '56.0', ')', 'add(', '\#0', '56.0', ')', 'add(', '\#1', '21.0', ')'] \\
    \bottomrule
    \end{tabular}%
  \caption{Case study. Type \uppercase\expandafter{\romannumeral1}: MT2Net produces the correct program length but takes a wrong operand. Type \uppercase\expandafter{\romannumeral2}: MT2Net under-generates the program sequence. Type \uppercase\expandafter{\romannumeral3}: MT2Net over-generates the program sequence. Type \uppercase\expandafter{\romannumeral4}: MT2Net selects the wrong operator and operands and ends the decoding early. Type \uppercase\expandafter{\romannumeral5}: MT2Net takes the wrong operand and is stuck in the loop.}
  \label{tab:5}%
\end{table*}%
We manually inspect a few samples of MT2Net and our approach from the development set for a case study. Results are shown in Table~\ref{tab:5}.

Table~\ref{tab:5} shows that our method can produce a more accurate program than MT2Net when the number of reasoning steps is correctly predicted (Type \uppercase\expandafter{\romannumeral1}). NAPG can predict the number of reasoning steps correctly (Type \uppercase\expandafter{\romannumeral2} and Type \uppercase\expandafter{\romannumeral3}), and generate the correct program sequences for the interpreting of the answers (Type \uppercase\expandafter{\romannumeral4} and Type \uppercase\expandafter{\romannumeral5}) when MT2Net fails. We conjecture that the good performance of NAPG might benefit from the weighted combination of individual loss functions (Eq.~\ref{eqa:tloss}), which allows us to tune the model for specific goals (as shown in Table~\ref{tab:lamda}), while the sequence generation of MT2Net fully relies on the single token prediction loss.

\section{Related Work}
\paragraph{Hybrid tabular-textual QA} \citet{chen2020hybridqa} present the first hybrid tabular-textual QA dataset, HybridQA, by linking table cells to Wiki pages via hyperlinks manually, and the answer is usually a span or paragraphs obtained from heterogeneous information.  \citet{zhu2021tat} and \citet{chen2021finqa} present TAT-QA and FinQA based on financial reports, which require numerical reasoning. TAT-HQA \citep{li2022learning} and ConvFinQA \citep{chen2022convfinqa} are extensions of these two datasets respectively. \citet{zhao2022multihiertt} present the MultiHiertt dataset, which contains multiple hierarchical tables and longer unstructured text. A more complex reasoning process across multiple tables and paragraphs is required to correctly answer the question.

\paragraph{Numerical Reasoning} Numerical reasoning ability is very important for many NLP tasks \citep{thawani2021representing,pal2021investigating}, especially in QA, such as text QA \citep{dua2019drop,ran2019numnet,hu2019multi,shao2021mutual,guo2021improving,zhang2021noahqa}, table QA \citep{iyyer2017search,herzig2020tapas,katsis2021ait,liu2021tapex,yang2022tableformer,pan2022end}, and hybrid tabular-textual QA \citep{chen2021finqa,zhu2021tat,li2021tsqa,li2022learning,zhao2022multihiertt,chen2022convfinqa,deng2022pacific}. \citet{geva2020injecting,berg2020empirical,pi2022reasoning} attempt to inject numerical reasoning ability into pre-trained language models.  \citet{zhu2021tat,li2022learning,zhu2022towards} perform a single arithmetic operation based on predefined operators. The encoder-decoder transformer such as T5 \citep{raffel2020exploring} or decoder transformer such as GPT-2 \citep{radford2019language} can be used to autoregressively decode program sequences, but previous work \citep{chen2022convfinqa} has verified that they do not have advantages in numerical reasoning. FinQANet \citep{chen2021finqa} and MT2Net \citep{zhao2022multihiertt} can perform better for multi-step reasoning, both of them use the LSTM decoder to autoregressively generate the program.

\section{Conclusion}

Hybrid tabular-textual question answering (QA) requires reasoning from heterogeneous information, and numerical reasoning is its key challenge compared to extractive QA. To address the severe exposure bias issue of current autoregressive methods when program generation performance is far from good, we present a non-autoregressive program generation (NAPG) framework for numerical reasoning, which facilitates program generation in parallel. Our framework independently generates complete program tuples containing both the operator and its operands. Compared to previous autoregressive decoding methods, NAPG does not suffer from exposure bias, and can significantly boost program generation speed.

Our experiments on the ConvFinQA and MultiHiertt datasets show that: 1) our proposed model can bring about large improvements over the strong FinQANet ($+5.06$/$+4.80$ Exe/Prog Acc points) and MT2Net ($+7.97$/$+6.38$ EM/F1 points) baselines, establishing the new state-of-the-art performance, while being much faster ($\sim$21x) in program generation. 2) the performance drop of our method is also significantly smaller than the autoregressive LSTM decoder of MT2Net with increasing numbers of numerical reasoning steps.

\section*{Limitations}

We believe that this paper provides useful insights for the numerical reasoning part in hybrid tabular-textual QA. However, there are limitations to our work:

\begin{itemize}
    \item We only study the program generation of MT2Net and FinQANet while keeping the other parts unchanged. That said, the numerical reasoning is the main challenge of the hybrid textual-tabular QA task compared to traditional extractive QA studies, and how to address the exposure bias issue of program generation is the main focus of this work.
    \item Our experiments are mainly conducted on the MultiHiertt dataset even though we also tested the effectiveness on the ConvFinQA dataset. However, MultiHiertt is the more challenging compared to the other benchmarks \citep{zhao2022multihiertt}, which does not involve either multiple tables and longer texts \citep{zhu2021tat,chen2021finqa,li2022learning} or multi-step numerical reasoning \citep{chen2020hybridqa,li2021tsqa}.
\end{itemize}

\if else
\section*{Ethics Statement}

\section*{Acknowledgements}
\fi

\bibliography{anthology,custom}
\bibliographystyle{acl_natbib}

\clearpage

\end{document}